\documentclass[conference]{IEEEtran}
\IEEEoverridecommandlockouts

\usepackage{cite}
\usepackage{amsmath,amssymb,amsfonts}
\usepackage{algorithmic}
\usepackage{graphicx}
\usepackage{textcomp}
\usepackage{xcolor}
\usepackage[caption=false]{subfig}

\usepackage{pgfplotstable} 
\usepackage{booktabs} 

\def\BibTeX{{\rm B\kern-.05em{\sc i\kern-.025em b}\kern-.08em
    T\kern-.1667em\lower.7ex\hbox{E}\kern-.125emX}}
\begin{document}

\definecolor{myyellow}{RGB}{255, 200, 100}
\definecolor{myred}{RGB}{255, 124, 124}
\definecolor{mygreen}{RGB}{131, 210, 122}

\setlength{\tabcolsep}{3pt}

\pgfplotsset{compat=1.16} 
\pgfplotstableset{
    col sep=comma, 
    string type,
    header=false,
    display columns/0/.style={
        column type = {r}
    },
    every head row/.style={
        output empty row
    },
    every row no 0/.style={
        before row=\toprule,
        after row=\midrule
    },
    every last row/.style={
        after row={\bottomrule}
    },
    every column/.code={
        \ifnum\pgfplotstablecol>0
            \pgfkeysalso{column type/.add={|}{}}
        \fi
    },
    highlightrow/.style={
        postproc cell content/.append code={
           \count1=\pgfplotstablerow
            \advance\count1 by1
            \ifnum\count1=#1
            \pgfkeysalso{@cell content/.add={\bfseries\scriptsize}{}}
            \fi
        },
    },
    highlightcol/.style={
        postproc cell content/.append code={
           \count0=\pgfplotstablecol
            \advance\count0 by1
            \ifnum\count0=#1
            \pgfkeysalso{@cell content/.add={\scriptsize}{}}
            \fi
        },
    },
    highlightcol={1},
    highlightrow={1}
}

\title{A Comparison of Few-Shot Learning Methods for Underwater Optical and Sonar Image Classification}

\author{
    \IEEEauthorblockN{1\textsuperscript{st} Mateusz Ochal}
\IEEEauthorblockA{\textit{Heriot-Watt University}\\
Edinburgh, UK \\
m.ochal@hw.ac.uk}
\and
\IEEEauthorblockN{2\textsuperscript{nd} Jose Vazquez}
\IEEEauthorblockA{\textit{SeeByte}\\
Edinburgh, UK \\
jose.vazquez@seebyte.com}
\and
\IEEEauthorblockN{3\textsuperscript{rd} Yvan Petillot}
\IEEEauthorblockA{\textit{Heriot-Watt University}\\
Edinburgh, UK \\
y.r.petillot@hw.ac.uk}
\and
\IEEEauthorblockN{4\textsuperscript{th} Sen Wang}
\IEEEauthorblockA{\textit{Heriot-Watt University}\\
Edinburgh, UK \\
s.wang@hw.ac.uk}
}

\maketitle

\begin{abstract}
Deep convolutional neural networks generally perform well in underwater object recognition tasks on both optical and sonar images. Many such methods require hundreds, if not thousands, of images per class to generalize well to unseen examples. However, obtaining and labeling sufficiently large volumes of data can be relatively costly and time-consuming, especially when observing rare objects or performing real-time operations. Few-Shot Learning (FSL) efforts have produced many promising methods to deal with low data availability. However, little attention has been given in the underwater domain, where the style of images poses additional challenges for object recognition algorithms. To the best of our knowledge, this is the first paper to evaluate and compare several supervised and semi-supervised Few-Shot Learning (FSL) methods using underwater optical and side-scan sonar imagery. Our results show that FSL methods offer a significant advantage over the traditional transfer learning methods that fine-tune pre-trained models. We hope that our work will help apply FSL to autonomous underwater systems and expand their learning capabilities.
\end{abstract}

\section{Introduction}\label{sec:introduction}
Underwater object recognition is generally more challenging than in the usual indoor/outdoor environments due to the unique interaction of light in the water that distorts optical images. Water molecules, dust, and other floating particles can cause substantial attenuation of light, limit sensing range, affect the color, and introduce haze into pictures. In addition to optical cameras, acoustic sensors equip many underwater systems. Unaffected by lighting conditions, acoustic sensors have an extended sensing range, offering a significant advantage. However, sonar still greatly suffers from noisy sensor input and lower resolution. These characteristics can negatively affect the performance of deep convolutional neural networks (DCNN) \cite{mohammed2017deep, rutledge2018intelligent}.

A key component for achieving good performance is training on large datasets \cite{LeCun2015dl}. However, obtaining larger datasets can be expensive and impractical in the marine setting due to the high operational costs and time constraints associated with underwater missions. A low abundance of some types of objects can further limit extensive gathering of data. Moreover, in real-time operations, it can be infeasible to perform rigorous labeling of data. Finding an algorithm capable of learning from only a handful of samples would be beneficial not only in the underwater domain but also to the general robotics and computer vision communities. 

A variety of regularisation techniques address the problem of learning with limited data. One of the popular methods is transfer learning (TL), which has seen overall success in the underwater setting \cite{mohammed2017deep}. In TL, a network is typically trained on a significantly larger but readily available dataset, and later the model is fine-tuned on a smaller domain-specific dataset. However, TL alone may still require thousands of images in the smaller dataset to generalize reliably.

Over the past several years, there has been a renewed effort in developing more efficient algorithms to perform \textit{Few-Shot Learning (FSL)}. FSL methods are commonly trained through meta-learning (e.g., MAML \cite{finn2017maml}) that aims to teach models how to learn from a few samples. Recent efforts have created a range of robust methods and proved to be promising for alleviating the problem of learning with limited data. 

While FSL methods have been extensively tested on generic classification datasets, little attention has been given to practical underwater scenarios. To this end, we compare various FSL methods on a range of challenging optical and sonar datasets. We identify the state-of-the-art and highlight some challenges still faced by FSL methods.
Our main contributions can be summarised as follows.
\begin{itemize}
    \item To the best of our knowledge, our work is the first to compare the performance of several FSL methods for underwater sonar and optical image classification.
    \item We show that FSL methods offer a significant advantage over the traditional methods of fine-tunning.
    \item We show that pre-meta-training FSL methods on general-purpose datasets can further improve performance, even when the image types differ significantly.
    \item We discuss the practicality of using FSL methods in realistic underwater robotics scenarios, highlighting their limitations, and proposing directions for future research.
\end{itemize}

This paper is structured as follows. We begin with an overview of the related literature in section~\ref{sec:related_works}, describing the current efforts of training deep learning models with limited data, few-shot learning, and on underwater images. In section~\ref{sec:datasets} we explain the datasets used for our experiments, before describing the examined methods in section~\ref{sec:methodology} and the experimental setup in section~\ref{sec:experiments}. We report results in section~\ref{sec:results}, and discuss the limitations of few-shot learning and our experiments in section~\ref{sec:discussion}.

\section{Related Work}\label{sec:related_works}
\subsection{Learning with Limited Data}
DCNN models can contain well into tens of millions of trainable parameters. As an example, EfficientNet-B7 \cite{mingxing2019efficient}, which achieves state-of-the-art performance on ImageNet \cite{russakovsky2015ilsvrc}, contains about 66M trainable parameters. Generally, the more parameters a model has, the greater its capacity to learn intricate patterns present in the data and achieve higher accuracy performance \cite{mingxing2019efficient}.

However, large models tend to overfit on small training datasets because they cannot learn a correct distribution of data due to the low variance of the training set, leading to high bias. The problem of overfitting has been addressed by numerous regularisation techniques, such as weight-decay \cite{plaut1986backprop, lang1990dimredu}, dropout \cite{hinton2012improving, labach2019dropout}, data augmentation \cite{simard2013best}, transfer learning \cite{pan2010transfer} and many others \cite{kukacka2017regularization}. A regularisation method can be \textit{``any supplementary technique that aims at making the model generalize better, i.e., produce better results on the test set"} \cite{kukacka2017regularization}.

\subsection{Few-Shot Learning}
Few-Shot Learning (FSL) models aim to classify between classes from only a handful of sample representatives. Specifically, in a $k$-shot $n$-way FSL classification task, a model is given a small training set (called a \textit{support set}) consisting of $n$ never-seen-before classes with $k$ image-label pairs per class. The goal is to use the support set to correctly classify a small evaluation set (called a \textit{target set}) containing a different set of image-labels pairs sampled the same $n$ classes. One-shot learning is an extreme case of FSL, which utilizes only a single support sample from each class ($k=1$).

The approaches to FSL algorithms can be broadly categorized into five categories \cite{chen2019closer}: metric-learning, optimization-based, hallucination, probabilistic, and domain adaptation. \textit{Metric-learning} approaches (such as Prototypical Networks \cite{snell2017protonets, ren2018kmeans, ayyad2019consistent}) learn a feature extractor function capable of uniquely describing images from novel classes. \textit{Optimization-based} approaches (such as MAML \cite{finn2017maml} and Meta-Learner LSTM \cite{ravi2017optimization}) aim to achieve efficient learning through a guided optimization process on the support set. \textit{Hallucination} or \textit{data augmentation} techniques perform affine and color transformations on the support set to create additional data points, for example, \cite{zhang2018metagan} exploits an imperfect Generative Adversarial Network to generate additional negative examples that refine the class boundaries in feature space. \textit{Probabilistic} methods use Bayesian inference to learn and classify samples (eg., GPShot \cite{patacchiola2019gpshot}). \textit{Domain adaptation} or \textit{transfer-learning} that are pre-trained using classical supervised learning; these include fine-tuning baselines as well as varients, like Baseline\texttt{++} \cite{chen2019closer}. 

FSL methods often learn through meta-learning, which employs three phases: meta-training, meta-validation, meta-testing\footnote{Due to clashing terminology of two communities, we make it explicit when referring to the meta-learning training/evaluation (by adding a prefix `meta-') as opposed to support-set learning and target set evaluation.}. During meta-training, models can learn general features and hyperparameters that can be used later in the FSL task. \textit{Episodic training} \cite{vinyals2016matching} is a popular way of meta-training where a learner model is repeatedly exposed to batches of FSL classification tasks sampled from a more extensive but different set of classes. This process allows methods to exploit readily available datasets such as ImageNet \cite{russakovsky2015ilsvrc}.

\subsection{Underwater Object Classification}

Underwater object classification faces many unique challenges. Optical images' quality is strongly affected by the interactions of light with water molecules and other floating particles. These interactions introduce haze, noise (blur and `marine snow' \cite{leonard2010underwater}), discoloring, and non-uniform illumination of objects. These factors can combine with various levels of strengths, and make object classification much more challenging to perform. Standard computer vision datasets (e.g., ImageNet) contain only up to a few underwater classes and do not generalize well to underwater optical datasets that contain higher levels of noise and color distortion \cite{mohammed2017deep,xu2018fish}. 

Additionally, there is a lower abundance of publically-available labeled underwater datasets. Many authors \cite{mohammed2017deep, rimavicius2017seafloor, xu2018fish, levy2018limited, tamou2018fish} choose to train neural networks using transfer learning, by pre-training on nonspecialist datasets such as ImageNet \cite{russakovsky2015ilsvrc}, and then fine-tuning last of few layers of the pre-trained model on the smaller underwater dataset. Some authors such as \cite{xu2018fish} apply rigorous data augmentation (including rotation, random cropping, flipping, and color-shifting), which further boosts performance. Some authors, such as \cite{yoon2015dehazing, sahu2014enhancement, berman2018color, lu2019restoration}, parse datasets using image enhancement methods to improve the quality of images by restoring the actual color of objects, remove haze, and denoise. Image enhancement techniques can aid human visibility, and some authors such as \cite{lu2019restoration} show them improving object tracking performance. 

Due to the limitation of optical vision, it is common to equip underwater vehicles with supplementary sonar cameras \cite{ferreira2016fusion}. Imaging sonar has become a widely adopted solution for providing measurements in many practical underwater operations \cite{lee2018sonar, rutledge2018intelligent, barngrover2015sonarmines, paull2014navigation}. It offers significant advantages over optical cameras due to its robustness to water turbidity and variable lighting conditions. Side-scan sonar (SSS) is particularity popular for surveying and mapping due to its wide coverage and bathymetric capabilities \cite{rutledge2018intelligent}. It can have a range of over a hundred meters. However, the acoustic signal is not perfect and has its limitations. For example, it does not provide any color information and has a lower resolution than optimal images taken with modern cameras. The resolution varies with the distance of detected objects, and there is a trade-off between accuracy and range. The random sensor noise, viewing angle dependency, and sonar reflection of materials further contribute to the difficulty of working with sonar. As a result, it is common to equip vehicles and take advantage of both sensors. Although fusing input signals from both sensory modalities is complex and uncommon, some successful attempts have been made \cite{ferreira2016fusion}.

Despite the challenging nature of sonar data, \cite{rutledge2018intelligent} has successfully applied a pre-trained ResNet-50 \cite{he2015residual} (on ImageNet \cite{russakovsky2015ilsvrc}) for a reliable shipwreck recognition system. \cite{lee2018sonar} has applied a Faster-RCNN \cite{ren2015fasterrcnn} with rigorous data augmentation for underwater object detection on both real and simulated sonar images. Transformantions on the training set included color inversion, horizontal and vertical flipping, scaling, rotation, and translation. 

In the context of few-shot learning, and to the best of our knowledge, only one research paper has applied FSL on underwater sonar images \cite{chen2019accustic_siemese}. However, the authors evaluate only a single method, called Siamese Networks \cite{koch2015siamese}, with no comparisons between alternative methods. Moreover, no quantifiable measure (such as accuracy) is reported offering limited insight into the underwater FSL problem.

\section{Datasets}\label{sec:datasets}
FSL models are typically meta-trained using three disjoint dataset splits, one for each meta-learning phase: meta-training, meta-validation, and meta-testing. Unlike in classical supervised learning, the classes for each phase are strictly non-overlapping. Mini-ImageNet \cite{ravi2017optimization} is a popular benchmarking dataset for FSL models. It is a downscaled subset of ImageNet-2012 \cite{russakovsky2015ilsvrc} containing only 100 of the original classes and only a few underwater classes and no sonar images. It is split into 64/16/20 classes for meta-training/meta-validation/meta-testing phases, respectively.

In our experiments, we evaluated methods on two color and two simulated-sonar datasets - offering an easier and a more difficult setting for each modality. When selecting datasets, we had to meet specific criteria, namely:
\begin{enumerate}
  \item the datasets had to be of underwater images to fit the scope of this research,
  \item contain at least 15 distinct classes to perform 5-way classification during each of the meta-learning phases,
  \item contain at least 40 images per class to fit the minimum support/target set setup.
\end{enumerate}
For these reasons, we chose the publically-available Fish Recognition dataset \cite{fisher2016fish} and a privately-held Pipeline Feature dataset containing higher levels of blur and discoloration. From the original 23 classes of the fish dataset, we filtered classes with less than 40 samples. The remaining 19 classes were divided into 9/5/5 classes for meta-training/meta-validation/meta-testing phases, respectively. Similarly, in the pipeline dataset, containing 16 classes in total, we used 6/5/5 classes. All images were scaled to 84 by 84 pixels. In contrast to the fish dataset, Pipeline Features is significantly color shifted towards the green end of the color spectrum, offering a more challenging but realistic underwater scenario. 

\begin{figure} 
    \centering
    \subfloat[Mini-ImageNet\cite{vinyals2016matching}, showing a wolf, dog, lipstick, ant, and some fish.]{
      \includegraphics[width=0.98\linewidth]{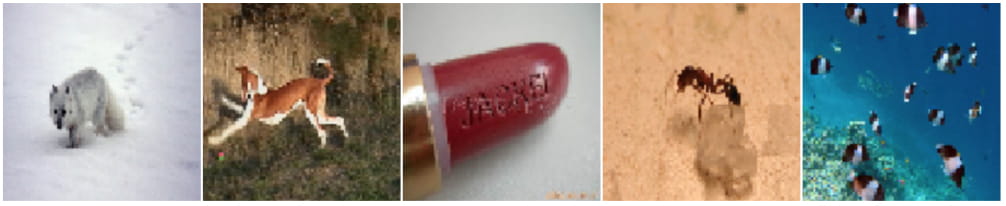}}
      \vspace{1mm}
    \subfloat[Fish Recognition\cite{fisher2016fish}, showing five different fish species.]{
      \includegraphics[width=0.98\linewidth]{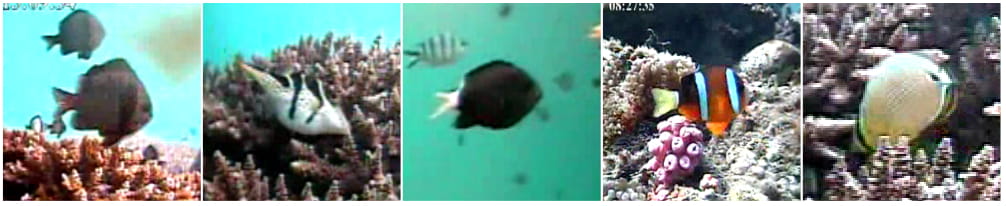}}
      \vspace{1mm}
    \subfloat[Pipeline Features include an anode, grout bag, shell, fish and sea urchin.]{
      \includegraphics[width=0.98\linewidth]{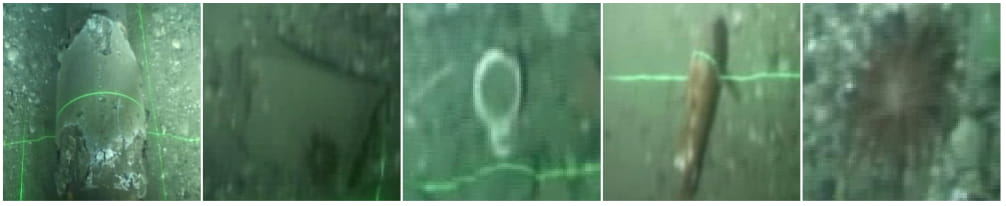}}
      \vspace{1mm}
    \subfloat[SSS (flat), showing an anchor, cube, plane, boat, and pyramid.]{
      \includegraphics[width=0.98\linewidth]{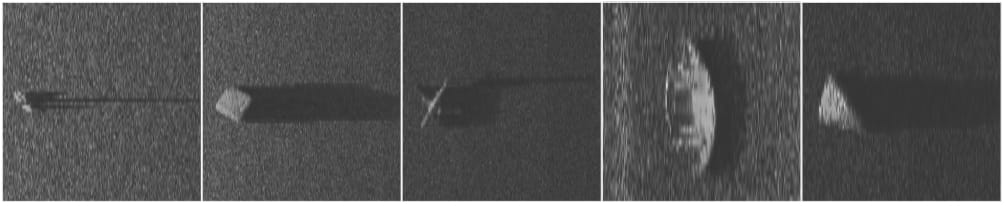}}
      \vspace{1mm}
    \subfloat[SSS (rippled), showing an anchor, cube, plane, boat, and pyramid.]{
      \includegraphics[width=0.98\linewidth]{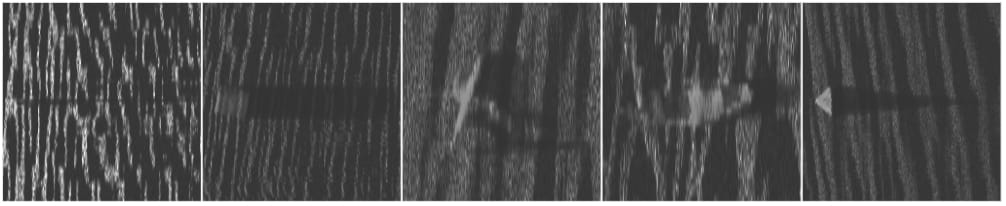}}
  \caption{Image examples from the datasets used in this work, representing only a small subset of available classes.}
  \label{fig:samples} 
\end{figure}

Sonar is integral in many underwater robotics systems. Due to the scarce availability of public sonar datasets, a specialized side-scan sonar (SSS) simulator was used to generate two datasets, as described in \cite{karjalainen2019gan}. The simulator works by ray tracing a 3D Computer-Aided Design (CAD) model to emulate the signal received by a sonar sensor, producing realistic shadows and highlights of synthetic contacts (objects). We note that the sonar simulator's quality was validated in experiments with human participants and DCNN networks, which were unable to distinguish between real and simulated imagery \cite{karjalainen2019gan}. We inserted 18 different synthetic contacts into two types of simulated seabeds at various orientations and depth levels. We refer to the two seabed types as \textit{flat} and \textit{rippled}, with the latter offering a more challenging scenario. For each type of seabed, we used 8/5/5 classes for the three meta-learning phases. To generate the images, we cropped an area centered around each object with a large margin around it to include the shadows. Each image was then scaled to 84 by 84 pixels. Figure~\ref{fig:samples} shows image examples.

In our experiments, we also evaluated a couple of semi-supervised FSL methods. To allow these methods to utilize unlabeled samples, we further partitioned each meta-learning dataset split, into a 40\%/60\% labeled/unlabeled partitions. 

\section{Methodology}\label{sec:methodology}
In this section, we give a low-level description of FSL methods evaluated in this work: Prototypical Network (PN) \cite{snell2017protonets} and its variants, Relation Networks \cite{sung2017relationnet}, Soft K-Means ProtoNets \cite{ren2018kmeans}, and Consistent Prototypical Networks \cite{ayyad2019consistent}. We begin by formally introducing the task of supervised FSL classification and the appropriate methods. Later, we introduce the semi-supervised FSL setting.

\subsection{Fully-supervised few-shot learning definition}
Consider the problem of a $k$-shot $n$-way classification task sampled from a dataset $\mathcal{D}$. A model is given a \emph{support set}, $\mathcal{S} = \left\{ (x_1, y_1), ... , (x_{s}, y_{s}) \right\} \sim \mathcal{D}$, containing $n$ unique classes with $k$ images per class ($|\mathcal{S}|=k \times n$). The goal of the model is to correctly classify a \emph{target set}, $\mathcal{T} = \left\{ (x_1, y_1), ... , (x_{t}, y_{t}) \right\}  \sim \mathcal{D}$, containing different samples from the same $n$ classes (i.e. $X(\mathcal{T}) \cap X(\mathcal{S}) = \emptyset$ and $Y(\mathcal{T}) \equiv Y(\mathcal{S})$). \emph{Episodic training} \cite{vinyals2016matching} is a popular way to meta-train FSL models, where models are exposed to mini-batches of $k$-shot $n$-way classification tasks sampled from a similar but disjoint dataset $\mathcal{D}_{train}$, where $\mathcal{D}_{train} \cap \mathcal{D} = \emptyset$.


\subsubsection{\textbf{Prototypical Network}}
\begin{figure}[tbhp]
    \centering
    \includegraphics[height=2.75cm]{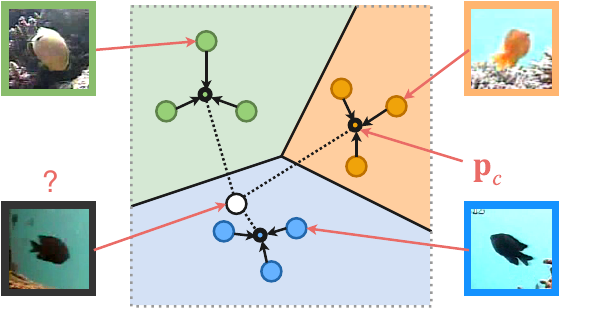}\caption{Prototypical Network. Prototypes $p_c$ are computed as the mean of the support samples belonging to a single class and mapped into an embedding space. A label for a target image is assigned based on the distances to the prototypes.} \label{fig:snell_prototypes}
    \vspace{0.75cm}
    \includegraphics[height=2.75cm]{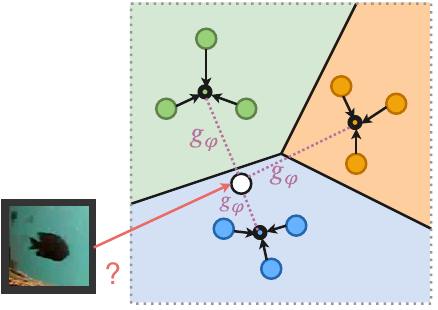}
    \caption[Relation Network]{Relation Network. Prototypes are computed in the same way as in Prototypical Networks. However, a label for a target image is assigned based on the score given by the relation module $g_\varphi$.}
    \vspace{0.75cm}
    \includegraphics[height=3cm]{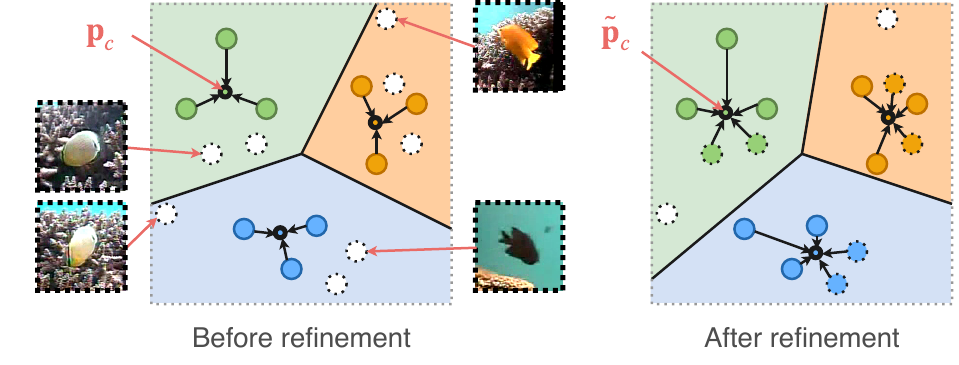}
    \caption{Prototypical Network with K-Means refinement. Information from unlabeled samples (marked with dashed outlines) is incorporated into the prototypes by a single iteration of Soft K-Means. Some samples are omitted in the process due to their low proximity to any prototype.}
    \vspace{0.75cm}
    \includegraphics[height=3cm]{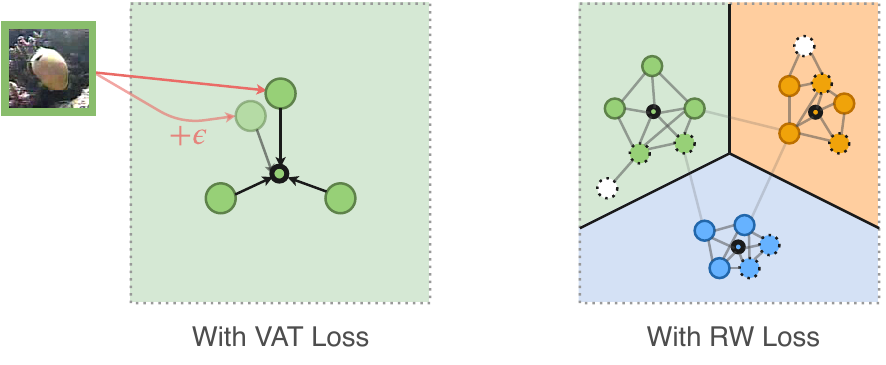}
    \caption{Consistent Prototypical Networks (CPN). CPN works on top of Prototypical Networks with and without the K-Means refinement. The Cross-Entropy loss is replaced by Virtual Adversarial Training loss (VAT) and Random-Walk loss (RW). During meta-training, VAT adds a small perturbation $\epsilon$ to each support sample before mapping it into the embedding space and calculating the prototypes. The goal of RW is to construct a tight neighborhood of samples for each class. The idea is that a `random walker' transverses similarity graphs between samples, and should rarely cross the natural class decision boundaries, thus, explicitly promoting clustering. }
    \label{fig:ren_refinement} 
\end{figure}

A Prototypical Network \cite{snell2017protonets} computes a representation of the support images for each class and assigns a class of a target image based on its similarity in embedding space. Specifically, support and target images are mapped into a feature space, through a non-linear mapping function $f_{\phi} : \mathbb{R} \rightarrow \mathbb{R}^M$, parameterized by the trainable parameters $\phi$. A class's prototype, $\textbf{p}_c \in \mathbb{R}^M$, is the mean of the mapped support samples belonging to a single class:
\begin{equation}\label{eq: prototype}
    \textbf{p}_c = \frac{\sum_i{f_\phi(x_i)z_{i,c}}}{\sum_i{z_{i,c}}}
\end{equation} 
where $z_{i,c} = \mathbf{1}$ when $y_i = c$ and $z_{i,c} = \mathbf{0}$ when $y_i \neq c$.
Given a target point $(x_j,y_j) \in \mathcal{T}$ and a distance function, $d: \mathbb{R}^M \times \mathbb{R}^M \rightarrow \left [ 0, + \infty \right )$, the model computes a similarity between the mapped target point and each of the prototypes. A softmax over the distances produces a probability distribution $p$ over the classes seen in the support set:

\begin{equation}\label{eq: targetset classification}
    p_\phi(y=c|x_j) = \frac{\text{exp}(-d( f_\phi(x_j), \textbf{p}_c ))}{\sum_{k'}{\text{exp}(-d(f_\phi(x_j),\textbf{p}_{c'}))}}
\end{equation}
The model is meta-trained by minimizing the average negative log-probability:
\begin{equation} \label{eq: log-prob loss}
    J(\phi) = -\text{log}\; p_\phi ( y = y_j | x_j )
\end{equation}
where $y_j$ is the true class of $x_j$. Figure~\ref{fig:snell_prototypes} shows an intuition of this method.

\subsubsection{\textbf{Relation Network}}
A Relation Network \cite{sung2017relationnet} augments the original Prototypical Network \cite{snell2017protonets} and replaces the distance measure, $d$, with a relation module $g_\varphi$, parametized by trainable parameters $\varphi$. Specifically, first mapped target points and the prototypes are combined with an operator $h(\textbf{p}_c, f_\phi(x_j))$ that concatenates each target point with each prototype. Secondly, each of the concatenated vectors are passed through the relation module to produce relation scores, $r_{k,j}$, between a class's prototype, $\textbf{p}_c$, and the target image $x_j$:
\begin{equation}\label{eq: relation score}
    r_{k,j} = \sum{x_i} g_\varphi \left(h\left(\textbf{p}_c, f_\phi(x_j) \right) \right)
\end{equation}
The embedding function $f_\phi$ and the relation module $g_\varphi$ are meta-trained end-to-end using the mean squared error (MSE). 

\subsection{Semi-supervised few-shot learning definition}
In a semi-supervised few-shot classification task, in addition to the labeled support-set, $\mathcal{S} \sim \mathcal{D}$, a model is also given an unlabeled set of images, $\mathcal{\tilde{S}} = \left\{ x_1, ... , x_{\tilde{s}} \right\}$, sampled from an unlabeled dataset $\mathcal{\tilde{D}}$. As before, the goal is to correctly classify the target set $\mathcal{T} \sim \mathcal{D}$. Episodic training \cite{vinyals2016matching} replaces datasets $\mathcal{D}$ and $\mathcal{\tilde{D}}$ with $\mathcal{D}_{train}$ and $\mathcal{\tilde{D}}_{train}$, respectively. Dataset $\mathcal{\tilde{D}}_{train}$ can be the same as $\mathcal{D}_{train}$, however, without losing generality we keep them seperate in notation.

\subsubsection{\textbf{Prototypical Network with K-Means Refinement} \cite{ren2018kmeans}}~\label{met: soft-kmeans-pn} This method also augments the original Prototypical Network \cite{snell2017protonets} and refines the prototypes using the unlabeled data $\mathcal{\tilde{S}}$. This method is almost identical to the original with the exception that the prototypes, $\textbf{p}_c$, are replaced by the \textit{refined prototype}, $\tilde{\textbf{p}}_c$, for each class, $k$. The refinement process use an iteration of the Soft K-Means algorithm (where $K=k$) on mapped images from $\mathcal{S}$ and $\mathcal{\tilde{S}}$.


The prototypes $\textbf{p}_c$ (defined in Eq.~\ref{eq: prototype}) act as the initial positions of the cluster centroids (i.e. $\tilde{\textbf{p}}_c \leftarrow \textbf{p}_c$). Each labeled example $x_i \in S^{(X)}$ is given a hard centroid assignment ($z_{i,c} = \mathbf{1}\left[y_i = c \right]$) since their label is considered known and therefore fixed. In contrast, each unlabeled sample $\tilde{x}_r$ is given a partitial (`soft') assignment $\tilde{z}_{r,c}$ to each cluster (of each class $k$) based on their Euclidean distance to the centroid locations. At each iterative step of the K-Means algorithm, the centroids are refined by integrating the adjusted assignments:
\begin{equation}
    \begin{split}
    \tilde{\textbf{p}}_c = \frac
{\sum_i{f_\phi(x_i)z_{i,c}}+\sum_r{f_\phi(\tilde{x}_r)\tilde{z}_{r,c}}}
{\sum_i{z_{i,c}}+\sum_r{\tilde{z}_{r,c}}}
,  \\ 
\text{where}\;\;\; 
\tilde{z}_{r,c} = \frac{\text{exp}(-d( f_\phi(\tilde{x}), \tilde{\textbf{c}}_c ))}{\sum_{c'}{\text{exp}(-d(f_\phi(\tilde{x}),\tilde{\textbf{c}}_{c'}))}} 
    \end{split}
\end{equation}
Although it is possible to perform multiple iterations of the clustering algorithm, the authors found that the performance does not improve after a single iteration.

\paragraph{{Soft K-Means PN + Cluster}}
The Soft K-Means approach described above assumes that $\mathcal{\tilde{S}}$ contains the same classes as $\mathcal{S}$, but this is unlikely to be true in a practical scenario. Classes that are not part of $\mathcal{S}$ are called \textit{distractors} since they are likely to interfere with the refinement process. To make the method more robust to distractors, the authors introduce an extra cluster ($K=k+1$) that acts as a `catch-all' cluster for anything that does not belong to the classes of interest, and thus, preventing any distractors from hindering with the refinement. The authors place the cluster at the origin ($\tilde{\textbf{p}}_c = \mathbf{0}$ for $c > n$) and introduce a learnable length-scale parameter, $q_c$, that reflects the amount of within-class variation. Thus, the partial assignment is defined as:
\begin{equation}
\begin{split}
    \tilde{z}_{r,k} = 
\frac{\text{exp}\left(
-\frac{1}{q_k^2}d\left(f_\phi\left(\tilde{x}\right), \tilde{\textbf{c}}_k \right)
-A(q_k)\right)}
{\sum_{k'}{\text{exp}\left(
-\frac{1}{q_k^2}d\left(f_\phi\left(\tilde{x}\right), \tilde{\textbf{c}}_{k'} \right)
-A(q_{k'})
\right)}} 
\\ 
\text{where}\;\;\;
A(q) = log(q) + \frac{1}{2}log(2\pi) 
\end{split}
\end{equation}
For simplicity, the authors set $q_{1...C}$ to $1$ in their experiments and only learn the length-scale of the distractor cluster $q_{n+1}$. Our experiments follow the same setup.

\paragraph{{Soft K-Means PN + Mask}}
The authors consider an alternative method to deal with distractor classes. Intuitively, a single distractor cluster is unlikely to not work well with higher numbers of distractor classes. To address these problems, instead of using a high-variance `catch-all' cluster, an image is labeled as a distractor if its embedding does not lie within legitimate proximity of any of the class' prototypes. Specifically, the Soft K-Means refinement process is altered as follows. Firstly, the normalized distances, $\tilde{d}$, are computed between examples $\tilde{x}_r \sim \mathcal{\tilde{S}}$ and prototypes $\textbf{p}_c$:
\begin{equation}
\begin{split}
    \tilde{d}_{r,c}= \frac{d_{r,c} }{\frac{1}{\tilde{M}}\sum_j{d_{r,c}} }
\end{split}
\end{equation}
where $d_{r,c}= d\left( f_\phi(x_r), \textbf{p}_c \right) = ||f({\tilde{x}_r) - \textbf{p}_c ||_2^2}$. Secondly, a small neural network computes learnable parameters $\beta_c$ and $\gamma_c$ from various statistics of the normalised distances (i.e. using the min, max, variance, skewness and kurtosis of $\tilde{d}_{r,c}$). The parameters $\beta_c$ and $\gamma_c$ help to establish how aggressively the unlabeled samples should influence centroids during the refinement process. The final refinement process of the \textit{Soft K-Means PN + Mask} method is:
\begin{equation}
    \begin{split}
    \tilde{\textbf{p}}_c = \frac
{\sum_i{f_\phi(x_i)z_{i,c}}+\sum_r{f_\phi(\tilde{x}_r)\tilde{z}_{r,c}m_{r,c}}}
{\sum_i{z_{i,c}}+\sum_r{\tilde{z}_{r,c}m_{r,c}}}
, \\
\text{where}\;\;\; 
m_{r,c} = \sigma \left( -\gamma_c\left(\tilde{d}_{r,c} - \beta_c \right )\right )
\end{split}
\end{equation}
where $m_{r,c}$ are the soft-masks computed by comparing the normalised distances to the learned thresholds.

\subsubsection{\textbf{Consitent Prototypical Network}} Consistent Prototypical Networks (CPNs) \cite{ayyad2019consistent} are also a semi-supervised FSL method capable of working with the original PN \cite{snell2017protonets} and the K-Mean refined PN \cite{ren2018kmeans}. The authors use virtual adversarial training (VAT) \cite{miyato2018vat}, and random walk (RW) loss \cite{kamnitsas2018rw, haeusser2017rw} to formulate a loss function that drives the meta-training process:
\begin{equation}
    \mathcal{L}_{SSL} =  \mathcal{L}_{VAT} + \mathcal{L}_{RW}
\end{equation}
Virtual adversarial training loss \cite{miyato2018vat} works on the assumption of \textit{local consistency}, also known as \textit{smoothness}, that two data points which are close together should get similar labels. In other words, if we add small perturbations to a point, it should not change its label by much. The local consistency loss of a point is calculated independently of the other points. Inspired by previous work \cite{kamnitsas2018rw, haeusser2017rw}, the authors of CPN introduce a \textit{global-consistency} loss that considers all data points and the overall structure of the embedding manifold. Let us consider points in the embedding space forming graph structures based on their similarity where the probability of going from a point to another varies based on the distance between the points. A loss can be calculated through a \textit{random-walk} over these similarity graphs constructed between unlabeled examples and the prototypes. The idea is that a random walker starting from a prototype should rarely cross the natural class decision boundaries, thus, explicitly promoting clustering. This can be achieved by allowing the random walker to take some fixed number of steps jumping between points, and maximizing the probability that the random walker gets back to the initial prototype within those steps.

\section{Experiments}\label{sec:experiments}
\subsection{Meta-training}\label{sec: exp metatrain}
For each dataset, three meta-training scenarios were constructed: meta-training on Mini-ImageNet \cite{ravi2017optimization}, meta-training on one of the underwater datasets, and meta-training using both datasets:
\begin{enumerate}
  \item \label{train_mini} \textbf{Meta-training on Mini-ImageNet}. In the first set of experiments, we meta-trained models on the meta-training split of Mini-ImageNet, following the original papers' setup. Specifically, ordinary PN \cite{snell2017protonets} was meta-trained using 5-shot 15-way classification tasks, while all the other methods used 5-shot 5-way tasks. Semi-supervised algorithms sampled additional 5 samples per class from the unlabeled partition of a relevant dataset split. All methods used 5 target images per class. The PN and Relation Networks were trained for $4 \times 10^5$ tasks but generally converged much sooner. Soft k-Means PNs were trained for $2 \times 10^6$ tasks but rarely improved beyond $5 \times 10^5$ tasks. CPNs were trained for $1.2 \times 10^6$ tasks. Evaluating the meta-testing split of Mini-ImageNet showed that our implementations achieved the within 3 accuracy points of the methods' claimed performances.

  \item \label{train_water} \textbf{Meta-training on an underwater dataset}. Similarly, we meta-trained the FSL models from random weight initialization on underwater datasets. We follow a similar setup as described above with a few notable changes to accommodate the smaller dataset sizes. Like other methods, ordinary PN \cite{snell2017protonets} was trained using 5-shot 5-way classification to accommodate the lower number of classes in the meta-training split. The ordinary PN and Relation Networks were trained using $4 \times 10^3$ tasks, but we found that the algorithms generally converged much sooner. Soft k-Means PNs and CPNs were trained for $5 \times 10^3$ tasks. 

  \item \label{train_both} \textbf{Meta-training on both datasets}. In this set of experiments, we pre-meta-trained the FSL models on the Mini-ImageNet dataset before meta-training on the underwater dataset. Specifically, we used the best meta-trained model on Mini-ImageNet (as described in point \ref{train_mini}), and we further meta-trained it on the underwater dataset (as described in point \ref{train_water} but without re-initialization).
\end{enumerate}

\subsection{Common evaluation and setup}
All experiments follow the same evaluation setup. That is, throughout the meta-training process, the models were meta-validated after every few-thousand tasks, and the best model was saved based on the performance on the meta-validation dataset split. At the end of meta-training, the best model was meta-tested on 1000 FSL 5-shot 5-way tasks sampled from the meta-testing split. During a task, models used the support set and the previously acquired knowledge to classify target set samples. We repeated each experiment 10 times for each algorithm, dataset, and meta-training type. Our results show the average target set accuracy. Due to the low number of classes in underwater datasets, we randomly picked resampled classes to be used for meta-training/meta-testing/meta-validation splits between each repeat. We reasoned that freezing the splits would create a bias towards specific FSL methods and create a skewed view of the methods' performance on the underwater dataset.

\subsection{Network Architectures}
All FSL models used a vanilla convolutional neural network consisting of 4 convolutional blocks. Each block was composed of a convolutional layer (each with 3 by 3 receptive fields, 64 filters, stride 1, and padding 0) followed by batch normalization \cite{ioffe2015batchnorm}, ReLU activation functions, and max-pooling. ConvNet baseline followed the same setup. Relation Networks used a relation network consisting of two convolutional blocks followed by a linear layer with one output.

\subsection{Fine-tuned Baselines}
In addition to few-shot learning methods, we selected a few fine-tune baselines for comparison. These include a range of convolutional networks that replace the processes of meta-training (as described in subsection~\ref{sec: exp metatrain}) with pre-training. The pre-training is performed on the meta-training split of a dataset and then fine-tuning the last layers on the support sets during the meta-testing phases.
\paragraph{\textbf{ConvNet}} We compared FSL methods with an equally powerful baseline model using the same 4 convolutional block architecture, and we called it ConvNet. The model contained an additional linear layer and a softmax over five output units (one for each class in the 5-way FSL task). The pre-training process was done over $4 \times 10^5$ mini-batches with batch size 64, and a learning rate of 0.001, slowly decaying at a rate of 0.9 after each $4 \times 10^4$ batches. After a few thousand mini-batches, the model was validated using FSL learning tasks, following the same meta-validation procedure as FSL models. Similarly, at the end of pre-training, the model was meta-tested on 1000 FSL 5-shot 5-way tasks sampled from the meta-testing dataset split. We performed fine-tuning by freezing all but the last linear layer of the baseline, which was randomly re-initialized, and fine-tuned on the support set ($25$ images for $5$-way $5$-shot task). The fine-tuning process performed 10 iterations with an initial learning rate of $0.01$, and a rapid decay rate of $0.5$ after each iteration. For each new evaluation task, we re-initialized the last layer with random weights. 
\paragraph{\textbf{ResNets}} Similarly to ConvNet, we pre-trained ResNet architectures. We used \textbf{ResNet-18} and \textbf{ResNet-50} trained from random weight initialisation. We also investigated versions of ResNets with a pre-trained set of weights that came with the PyTorch library, obtained from training on full-resolution ImageNet. We refer to these variants as \textbf{Initialised ResNet-18} and \textbf{Initialised ResNet-50}. To accommodate the smaller images size 84 by 84 pixels in the ResNet architecture, we automatically turned off max-pooling layers. Like the ConvNet, we pre-trained the four ResNet baselines and then fine-tuned the models' last layers on the support set. We used a learning rate of 0.0001 to accommodate the higher number of trainable parameters.

\begin{figure}[tbhp]
    \centering
    \includegraphics[width=0.97\linewidth]{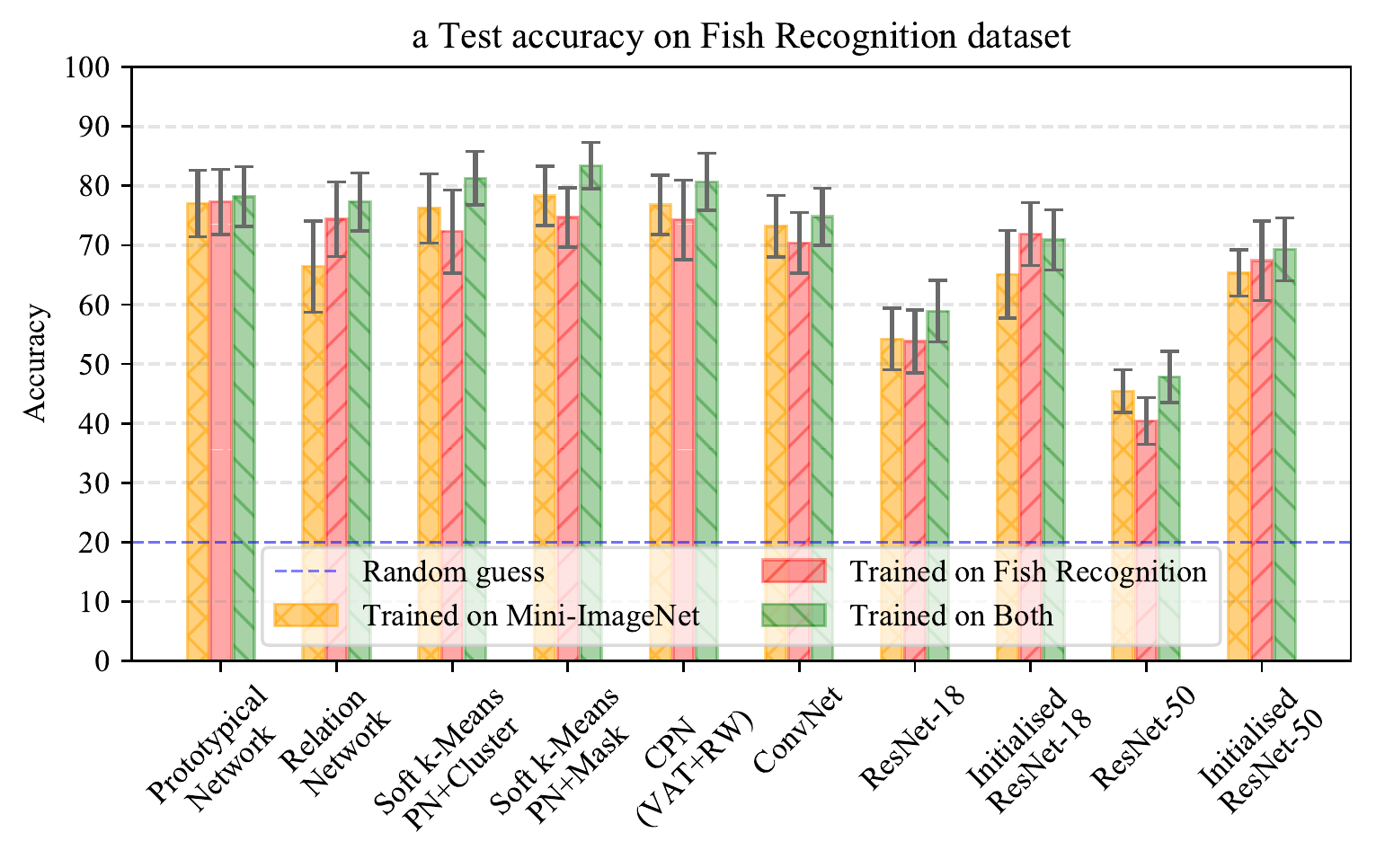}
    \includegraphics[width=0.97\linewidth]{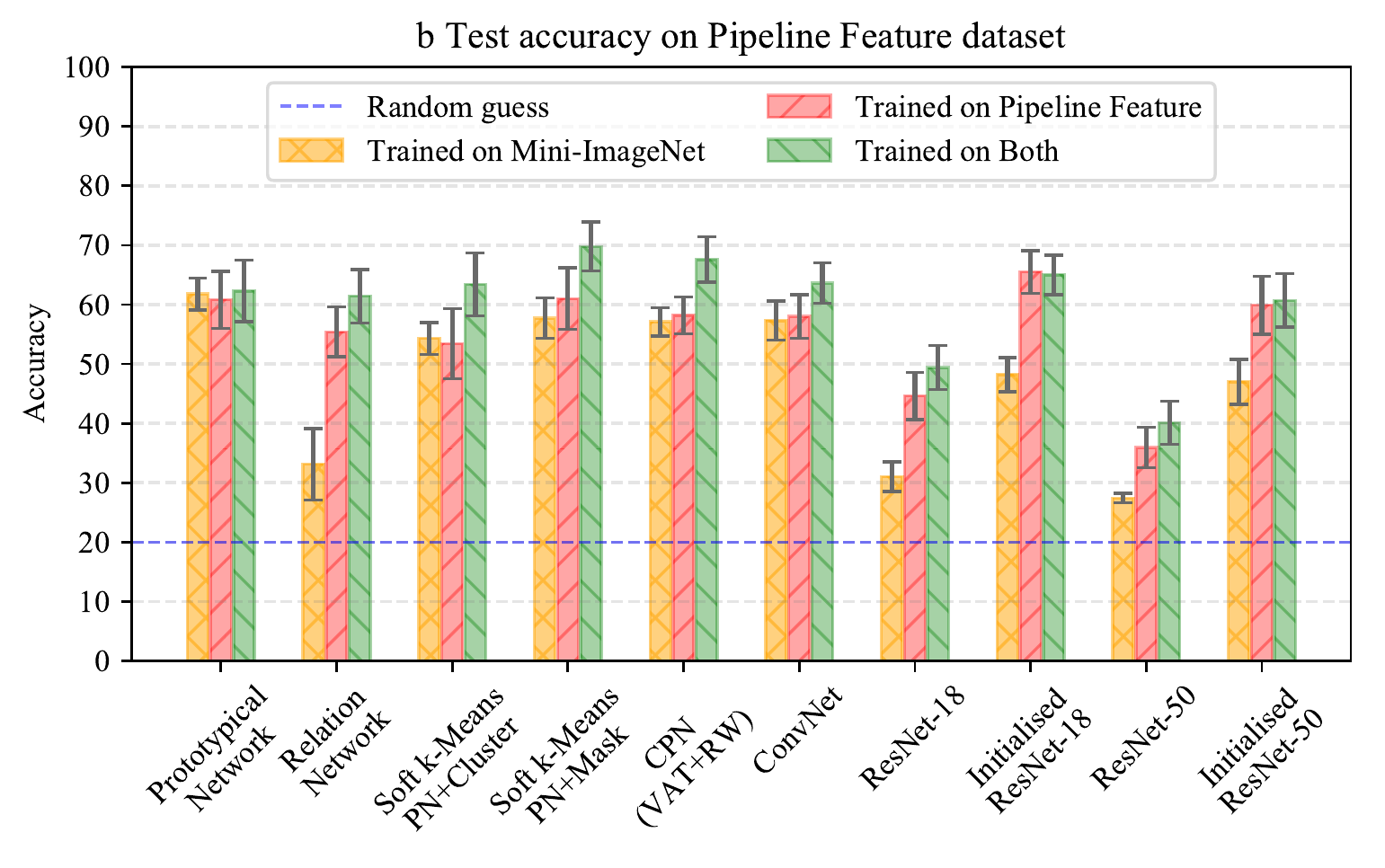}
    \includegraphics[width=0.97\linewidth]{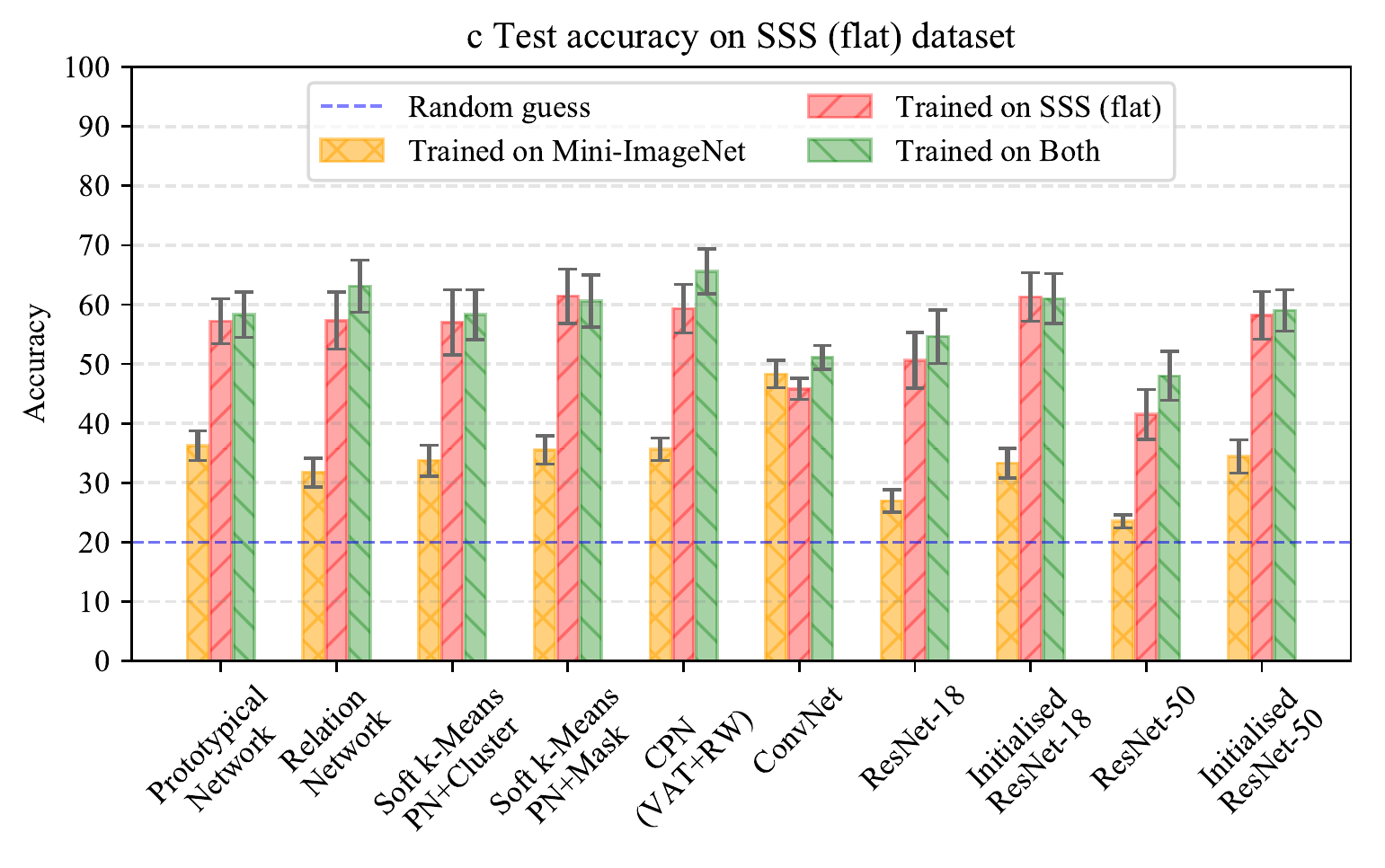}
    \includegraphics[width=0.97\linewidth]{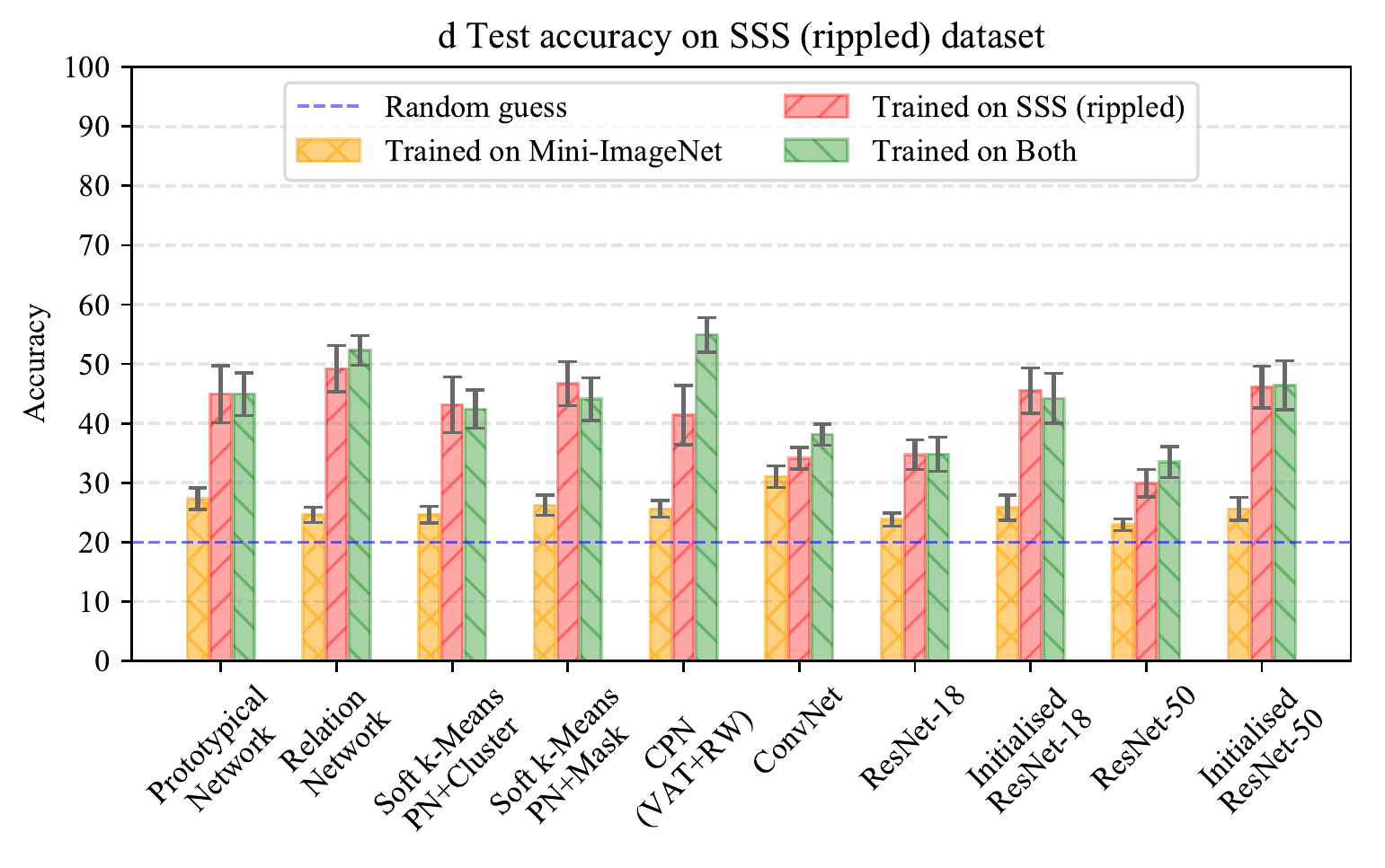}
    \caption{Test accuracy of models on the meta-testing split of the underwater datasets after meta-/pre- training models on the meta-training split of Mini-ImageNet (\textcolor{myyellow}{yellow}), the underwater dataset (\textcolor{myred}{red}), and both datasets (\textcolor{mygreen}{green}). The error bars show a 95\% confidence interval.}
    \label{fig:results}
\end{figure}

\begin{table}
\caption{Accuracy on testing split of Fish Recognition dataset after meta-/pre- training models on the meta-training split of Mini-ImageNet, Fish Recognition, and both datasets.}\label{tbl:fish}
\centering
\begin{filecontents*}{figs/tables_csv/fish-recognition-full-combined.csv}
Method,Mini-ImageNet,Fish Recognition,Both
Prototypical Network,77.0\textcolor{gray}{{\scriptsize ($\pm$5.6)}},77.3\textcolor{gray}{{\scriptsize ($\pm$5.5)}},78.2\textcolor{gray}{{\scriptsize ($\pm$5.0)}}
Relation Network,66.4\textcolor{gray}{{\scriptsize ($\pm$7.7)}},74.4\textcolor{gray}{{\scriptsize ($\pm$6.3)}},77.3\textcolor{gray}{{\scriptsize ($\pm$4.9)}}
Soft k-Means PN+Cluster,76.2\textcolor{gray}{{\scriptsize ($\pm$5.8)}},72.3\textcolor{gray}{{\scriptsize ($\pm$7.0)}},81.3\textcolor{gray}{{\scriptsize ($\pm$4.5)}}
Soft k-Means PN+Mask,78.3\textcolor{gray}{{\scriptsize ($\pm$5.0)}},74.7\textcolor{gray}{{\scriptsize ($\pm$5.0)}},83.4\textcolor{gray}{{\scriptsize ($\pm$3.9)}}
CPN (VAT+RW),76.8\textcolor{gray}{{\scriptsize ($\pm$5.0)}},74.3\textcolor{gray}{{\scriptsize ($\pm$6.7)}},80.7\textcolor{gray}{{\scriptsize ($\pm$4.8)}}
ConvNet,73.2\textcolor{gray}{{\scriptsize ($\pm$5.2)}},70.4\textcolor{gray}{{\scriptsize ($\pm$5.1)}},74.8\textcolor{gray}{{\scriptsize ($\pm$4.8)}}
ResNet-18,54.2\textcolor{gray}{{\scriptsize ($\pm$5.2)}},53.8\textcolor{gray}{{\scriptsize ($\pm$5.3)}},58.9\textcolor{gray}{{\scriptsize ($\pm$5.2)}}
Initialised ResNet-18,65.1\textcolor{gray}{{\scriptsize ($\pm$7.4)}},71.9\textcolor{gray}{{\scriptsize ($\pm$5.3)}},70.9\textcolor{gray}{{\scriptsize ($\pm$5.1)}}
ResNet-50,45.4\textcolor{gray}{{\scriptsize ($\pm$3.6)}},40.4\textcolor{gray}{{\scriptsize ($\pm$3.9)}},47.8\textcolor{gray}{{\scriptsize ($\pm$4.3)}}
Initialised ResNet-50,65.3\textcolor{gray}{{\scriptsize ($\pm$3.9)}},67.4\textcolor{gray}{{\scriptsize ($\pm$6.7)}},69.3\textcolor{gray}{{\scriptsize ($\pm$5.3)}}
\end{filecontents*}
\pgfplotstabletypeset[
    every row 4 column 1/.style={
      postproc cell content/.style={
        @cell content/.add={\bfseries}{}
      }
    },
    every row 1 column 2/.style={
        postproc cell content/.style={
          @cell content/.add={\bfseries}{}
        }
    },
    every row 4 column 3/.style={
      postproc cell content/.style={
        @cell content/.add={\bfseries}{}
      }
    },
  ]{figs/tables_csv/fish-recognition-full-combined.csv}
\vspace{0.4cm}
\caption{Accuracy on testing split of Pipeline Feature dataset after meta-/pre- training models on the meta-training split of Mini-ImageNet, Pipeline Feature, and both datasets.}\label{tbl:pipe}
\begin{filecontents*}{figs/tables_csv/pipeline-full-combined.csv}
Method,Mini-ImageNet,Pipeline Feature,Both
Prototypical Network,61.8\textcolor{gray}{{\scriptsize ($\pm$2.7)}},60.8\textcolor{gray}{{\scriptsize ($\pm$4.8)}},62.3\textcolor{gray}{{\scriptsize ($\pm$5.2)}}
Relation Network,33.1\textcolor{gray}{{\scriptsize ($\pm$6.0)}},55.4\textcolor{gray}{{\scriptsize ($\pm$4.2)}},61.4\textcolor{gray}{{\scriptsize ($\pm$4.5)}}
Soft k-Means PN + Cluster,54.3\textcolor{gray}{{\scriptsize ($\pm$2.7)}},53.4\textcolor{gray}{{\scriptsize ($\pm$5.9)}},63.4\textcolor{gray}{{\scriptsize ($\pm$5.3)}}
Soft k-Means PN + Masking,57.7\textcolor{gray}{{\scriptsize ($\pm$3.4)}},61.0\textcolor{gray}{{\scriptsize ($\pm$5.2)}},69.8\textcolor{gray}{{\scriptsize ($\pm$4.1)}}
CPN (VAT+RW),57.1\textcolor{gray}{{\scriptsize ($\pm$2.4)}},58.2\textcolor{gray}{{\scriptsize ($\pm$3.1)}},67.6\textcolor{gray}{{\scriptsize ($\pm$3.8)}}
ConvNet,57.3\textcolor{gray}{{\scriptsize ($\pm$3.3)}},58.0\textcolor{gray}{{\scriptsize ($\pm$3.7)}},63.6\textcolor{gray}{{\scriptsize ($\pm$3.4)}}
ResNet-18,31.0\textcolor{gray}{{\scriptsize ($\pm$2.5)}},44.6\textcolor{gray}{{\scriptsize ($\pm$4.0)}},49.4\textcolor{gray}{{\scriptsize ($\pm$3.7)}}
Initialised ResNet-18,48.2\textcolor{gray}{{\scriptsize ($\pm$2.9)}},65.5\textcolor{gray}{{\scriptsize ($\pm$3.6)}},65.0\textcolor{gray}{{\scriptsize ($\pm$3.3)}}
ResNet-50,27.4\textcolor{gray}{{\scriptsize ($\pm$0.8)}},35.9\textcolor{gray}{{\scriptsize ($\pm$3.4)}},40.1\textcolor{gray}{{\scriptsize ($\pm$3.6)}}
Initialised ResNet-50,47.0\textcolor{gray}{{\scriptsize ($\pm$3.8)}},59.9\textcolor{gray}{{\scriptsize ($\pm$4.9)}},60.7\textcolor{gray}{{\scriptsize ($\pm$4.5)}}
\end{filecontents*}
\pgfplotstabletypeset[
    every row 1 column 1/.style={
      postproc cell content/.style={
        @cell content/.add={\bfseries}{}
      }
    },
    every row 8 column 2/.style={
        postproc cell content/.style={
          @cell content/.add={\bfseries}{}
        }
    },
    every row 4 column 3/.style={
      postproc cell content/.style={
        @cell content/.add={\bfseries}{}
      }
    },
    ]{figs/tables_csv/pipeline-full-combined.csv}
\vspace{0.4cm}
\caption{Accuracy on testing split of SSS (flat) dataset after meta-/pre- training models on the meta-training split of Mini-ImageNet, SSS (flat), and both datasets.}\label{tbl:sss-flat}
\begin{filecontents*}{figs/tables_csv/simulated-sss-easy-combined.csv}
Method,Mini-ImageNet,SSS (flat),Both
Prototypical Network,36.2\textcolor{gray}{{\scriptsize ($\pm$2.5)}},57.2\textcolor{gray}{{\scriptsize ($\pm$3.8)}},58.3\textcolor{gray}{{\scriptsize ($\pm$3.8)}}
Relation Network,31.7\textcolor{gray}{{\scriptsize ($\pm$2.4)}},57.3\textcolor{gray}{{\scriptsize ($\pm$4.8)}},63.1\textcolor{gray}{{\scriptsize ($\pm$4.4)}}
Soft k-Means PN + Cluster,33.7\textcolor{gray}{{\scriptsize ($\pm$2.6)}},57.0\textcolor{gray}{{\scriptsize ($\pm$5.5)}},58.3\textcolor{gray}{{\scriptsize ($\pm$4.2)}}
Soft k-Means PN + Masking,35.5\textcolor{gray}{{\scriptsize ($\pm$2.4)}},61.4\textcolor{gray}{{\scriptsize ($\pm$4.6)}},60.6\textcolor{gray}{{\scriptsize ($\pm$4.4)}}
CPN (VAT+RW),35.6\textcolor{gray}{{\scriptsize ($\pm$1.9)}},59.3\textcolor{gray}{{\scriptsize ($\pm$4.1)}},65.6\textcolor{gray}{{\scriptsize ($\pm$3.8)}}
ConvNet,48.3\textcolor{gray}{{\scriptsize ($\pm$2.3)}},45.8\textcolor{gray}{{\scriptsize ($\pm$1.8)}},51.1\textcolor{gray}{{\scriptsize ($\pm$2.0)}}
ResNet-18,26.9\textcolor{gray}{{\scriptsize ($\pm$1.9)}},50.6\textcolor{gray}{{\scriptsize ($\pm$4.7)}},54.6\textcolor{gray}{{\scriptsize ($\pm$4.5)}}
Initialised ResNet-18,33.3\textcolor{gray}{{\scriptsize ($\pm$2.5)}},61.3\textcolor{gray}{{\scriptsize ($\pm$4.1)}},61.0\textcolor{gray}{{\scriptsize ($\pm$4.2)}}
ResNet-50,23.5\textcolor{gray}{{\scriptsize ($\pm$1.1)}},41.5\textcolor{gray}{{\scriptsize ($\pm$4.2)}},48.0\textcolor{gray}{{\scriptsize ($\pm$4.1)}}
Initialised ResNet-50,34.4\textcolor{gray}{{\scriptsize ($\pm$2.8)}},58.2\textcolor{gray}{{\scriptsize ($\pm$4.0)}},59.0\textcolor{gray}{{\scriptsize ($\pm$3.5)}}
\end{filecontents*}
\pgfplotstabletypeset[
    every row 6 column 1/.style={
      postproc cell content/.style={
        @cell content/.add={\bfseries}{}
      }
    },
    every row 4 column 2/.style={
        postproc cell content/.style={
          @cell content/.add={\bfseries}{}
        }
    },
    every row 5 column 3/.style={
      postproc cell content/.style={
        @cell content/.add={\bfseries}{}
      }
    },]{figs/tables_csv/simulated-sss-easy-combined.csv}
\vspace{0.4cm}
\caption{Accuracy on testing split of SSS (rippled) dataset after meta-/pre- training models on the meta-training split of Mini-ImageNet, SSS (rippled), and both datasets.}\label{tbl:sss-rippled}
\begin{filecontents*}{figs/tables_csv/simulated-sss-medium-combined.csv}
Method,Mini-ImageNet,SSS (rippled),Both
Prototypical Network,27.3\textcolor{gray}{{\scriptsize ($\pm$1.8)}},44.9\textcolor{gray}{{\scriptsize ($\pm$4.8)}},44.9\textcolor{gray}{{\scriptsize ($\pm$3.6)}}
Relation Network,24.6\textcolor{gray}{{\scriptsize ($\pm$1.3)}},49.2\textcolor{gray}{{\scriptsize ($\pm$3.9)}},52.3\textcolor{gray}{{\scriptsize ($\pm$2.5)}}
Soft k-Means PN + Cluster,24.6\textcolor{gray}{{\scriptsize ($\pm$1.4)}},43.1\textcolor{gray}{{\scriptsize ($\pm$4.7)}},42.4\textcolor{gray}{{\scriptsize ($\pm$3.2)}}
Soft k-Means PN + Masking,26.2\textcolor{gray}{{\scriptsize ($\pm$1.7)}},46.7\textcolor{gray}{{\scriptsize ($\pm$3.7)}},44.1\textcolor{gray}{{\scriptsize ($\pm$3.6)}}
CPN (VAT+RW),25.6\textcolor{gray}{{\scriptsize ($\pm$1.4)}},41.4\textcolor{gray}{{\scriptsize ($\pm$5.0)}},54.9\textcolor{gray}{{\scriptsize ($\pm$2.9)}}
ConvNet,31.0\textcolor{gray}{{\scriptsize ($\pm$1.8)}},34.1\textcolor{gray}{{\scriptsize ($\pm$1.8)}},38.1\textcolor{gray}{{\scriptsize ($\pm$1.8)}}
ResNet-18,23.8\textcolor{gray}{{\scriptsize ($\pm$1.1)}},34.7\textcolor{gray}{{\scriptsize ($\pm$2.5)}},34.8\textcolor{gray}{{\scriptsize ($\pm$2.9)}}
Initialised ResNet-18,25.8\textcolor{gray}{{\scriptsize ($\pm$2.1)}},45.5\textcolor{gray}{{\scriptsize ($\pm$3.8)}},44.2\textcolor{gray}{{\scriptsize ($\pm$4.2)}}
ResNet-50,22.9\textcolor{gray}{{\scriptsize ($\pm$1.0)}},29.9\textcolor{gray}{{\scriptsize ($\pm$2.3)}},33.5\textcolor{gray}{{\scriptsize ($\pm$2.6)}}
Initialised ResNet-50,25.6\textcolor{gray}{{\scriptsize ($\pm$1.9)}},46.1\textcolor{gray}{{\scriptsize ($\pm$3.5)}},46.4\textcolor{gray}{{\scriptsize ($\pm$4.1)}}
\end{filecontents*}
\pgfplotstabletypeset[
    every row 6 column 1/.style={
      postproc cell content/.style={
        @cell content/.add={\bfseries}{}
      }
    },
    every row 2 column 2/.style={
        postproc cell content/.style={
          @cell content/.add={\bfseries}{}
        }
    },
    every row 5 column 3/.style={
      postproc cell content/.style={
        @cell content/.add={\bfseries}{}
      }
    },]{figs/tables_csv/simulated-sss-medium-combined.csv}
\end{table}

\section{Results}\label{sec:results}
The results are presented in Tables~\ref{tbl:fish}-\ref{tbl:sss-rippled}, and Figure~\ref{fig:results}. Our experiments aim to answer the following questions:
\begin{itemize}
  \item Does meta-training on a general-purpose dataset generalize to underwater datasets? 
  \item Is there any advantage in pre-meta-training?
  \item Do FSL methods offer any advantage over traditional fine-tuning methods?
  \item What is state-of-the-art on underwater optical and sonar datasets?
\end{itemize}

For extra clarity in places, we will refer to the three meta-training scenarios by the numbers given in section~\ref{sec: exp metatrain}. 

\subsection{Generalization of Mini-ImageNet-trained models}
FSL models meta-trained on Mini-ImageNet alone (scenario \#1) achieved an average of 75.0\% and 52.8\% accuracy on Fish Recognition and Pipeline Features, respectively. On flat and rippled seabed sonar datasets, they achieved an average of 34.5\% and 25.7\% accuracy, respectively. These results confirm that the more the test dataset's style deviates from Mini-ImageNet, the worse the generalization of Mini-ImageNet trained models. 

Compared to the other two meta-training scenarios (scenario \#2: training on underwater datasets; and scenario \#3: training on both dataset), we observe an overall difference of -11\% and -16\% between scenario \#1 and \#2, and scenario \#1 and \#3, respectively - the only situation where scenario \#1 does better than scenario \#2 is on the Fish Recognition dataset. The better performance can be attributed to the similarity of Mini-ImageNet to the Fish Recognition dataset as well as the increased number of samples and classes in the Mini-ImageNet, which allowed the models to learn more generalizable features and achieve an overall higher accuracy. 

Furthermore, we observe that scenario \#1 models generalize poorly to sonar, with an average difference of -21.2\% accuracy points compared to the other two meta-training scenarios. In the more difficult sonar setting (with rippled seabed), the average model performance is only slightly better than random, reflecting the challenges caused by the significant style shift between optical and sonar images. It shows that general-purpose datasets alone are insufficient to meta-train few-shot learning models where the style of images differs significantly from the meta-testing split.

\subsection{Advantages of pre-meta-training}
For 17 out of 20 settings across all five FSL methods and four datasets, we observe that it is at least as good to meta-train models on both datasets (scenario \#3) as training with either of the other two scenarios. Overall, we observe an average improvement of 3.9\% accuracy points over the other two scenarios' best models - an average advantage of 5.2\% for optical images and 2.7\% for sonar images. Across the meta-training scenarios, we observe an average improvement of 16.5\% and 4.5\% over equivalent methods from scenario \#1 and \#2, respectively. This result demonstrates that there can be many gains of pre-meta-training on readily available datasets, even if the style of images differs from that of the meta-evaluation target set - a similar trend is observed in classical transfer learning approaches \cite{tamou2018fish}.

Interestingly, despite the significant differences in image style, the most substantial improvement of pre-meta-training (scenario \#3) can be observed for CPN on SSS (rippled) with 13.5\% improvement over the best of other two training scenarios. Although we observed poor generalization of Mini-ImageNet trained models in scenario \#1, the results of scenario \#3 show that some high-level features are still useful and can be utilized successfully during a later meta-training phase. 

In this study, we explored one way of meta-training on both datasets; meta-training once on ImageNet, then meta-training again on a specialized dataset. However, mixing datasets into a single dataset could be another way of combining them. However, we leave this investigation for future work.

\subsection{Advantages of Few-Shot Learning methods}
Comparing FSL methods with fine-tuned baselines, across all datasets and meta-/pre- training scenarios, we observe that in 10 out of 12 settings, there is at least one few-shot learning model that achieves at least as good performance as the equally powerful ConvNet baseline. We observe an average improvement of 7\% accuracy points using FSL methods over the ConvNet baseline, with up to 8.6\% on optical datasets and up to 16.8\% on sonar.

The FSL models can even achieve at least as good performance as the more powerful ResNet-18 and ResNet-50 baselines in 11 out of 12 settings. Interestingly, the pre-trained ResNet-50 (from random initialization) sometimes performed worse than the less powerful ResNet-18, which could be due to overfitting caused by the increased number of trainable parameters. The Initialized ResNets (pre-trained on full-scale ImageNet) performed overall the best out the baseline models.

On sonar datasets in scenario \#1, the ConvNet baseline does the best out of all of the methods, outperforming the FSL methods and the more powerful ResNet baselines. We theorize that this superior performance could be due to the adaptation ability of the fine-tuning process. Adjusting the network's weights using the support set has some advantage over the non-tunable meta-evaluation process of the Prototypical Networks and variants. It could be interesting to investigate optimization-based FSL methods; however, we leave this for future work. Although the more powerful ResNet baselines also performed fine-tuning, their performance was inferior to ConvNet on the sonar datasets in scenario \#1. It is likely that the ResNet networks, which contain many more convolutional layers, learn color-dependent features early in the network, which may impede the process of fine-tuning on sonar images. In contrast, ConvNet is a much shallower network, and the final network layer is more likely to contain high-level features that can easily be fine-tuned to the style of sonar images.

In some experiments, ResNet baselines do better than the FSL baselines. However, these models should not be directly compared since the underlying architecture of FSL methods is similar to the ConvNet that contains less trainable parameters and has a shallower architecture. For example, we found that ConvNet contained $1.3 \times 10^5$ trainable parameters, whereas were $1.2 \times 10^7$ parameters in ResNet-18 and $2.6 \times 10^7$ in ResNet-50, which is at two orders of magnitude greater. It would be interesting to substitute FSL models with a more powerful architecture. Work by \cite{oreshkin2018tadam} shows that using a more powerful backbone model in Prototypical Network significantly improves its performance. However, we leave this investigation to future work.

\subsection{State-of-the-art FSL on underwater datasets}
Soft K-Means PN achieves the best performance on Fish Recognition and Pipeline Feature, with 83.4\% and 69.8\% accuracy in scenario \#3, respectively - an improvement of 8.6\% and 6.8\% points over the ConvNet baseline model. On sonar datasets, CPN achieves the best performance with 65.6\% and 54.9\% accuracy on flat and rippled seabed in scenario \#3 - offering 14.5\% and 16.8\% point improvement over the ConvNet baseline. 



Generally, when meta-trained on both datasets, best semi-supervised methods tend to do slightly better than the best fully supervised FSL methods on the same dataset, with an average improvement of 6.4\% accuracy points on optical and 2.5\% on sonar, across all three meta-training scenarios. Interestingly, semi-supervised methods achieved better performance to fully supervised methods even though they only used 40\% of the labels. Their advantage could be due to at least two factors. On the one hand, the presence of 5 additional unlabeled samples per class exposes the algorithm to more information, which it could utilize when learning about new classes. On the other, the prototype refinement process could result in a more accurate representation of a classes' mean. Our supplementary experiments, on Soft K-Means PN models with no additional unlabeled samples, suggest that most of the performance gain is attributed to the post-processing of feature vectors, rather than the existence of additional data. In some experiments, additional data produced worse performance. However, more experiments would need to be collected to offer a more thorough insight, and we leave this investigation for future work. 

\section{Discussion}\label{sec:discussion}
Throughout the previous sections, we have seen FSL methods performing well on underwater optical and sonar images. In this section, we would like to highlight some limitations of our results as well as FSL methods in general that require further consideration before applying these methods on real-world robotics applications.

Few-shot learning methods work under a strict set of assumptions that might make them challenging to apply to practical settings. Firstly, achieving strong performance significantly depends on the choice of the support set. In some of our experiments, we found that the support set's choice was essential for capturing the intraclass differences. Moreover, FSL benchmarks typically assume that the support set is sampled uniformly from a single distribution. However, in real-world applications, the support set is likely to become available incrementally over time, contain a varying number of samples per class, and come from a highly correlated video frame stream. 

Moreover, in this work, the FSL methods were examined on a constrained classification problem where objects were present in the center of images. In a practical situation, the FSL classification models are likely to function on top of automatic target recognition (ATR) systems. The ATR system is likely to output a range of regions with various scales and objects placed anywhere within. Further considerations are required to apply FSL to work with ATR systems. 

Finally, the FSL methods examined assume the accessibility of all $k \times n$ images at once, with no future updates. In parallel work, we already investigate FSL methods in a general continual learning setting where the algorithms are exposed to new samples a small batch at a time \cite{antoniou2020cfsl}. However, more consideration is needed for learning with underwater images, as reflected in our experiments. 

\section{Conclusion}
In this work, we investigated few-shot learning (FSL) methods on four underwater datasets. For each method, we compared three meta-training scenarios: meta-training on a general-purpose dataset (Mini-ImageNet), on an underwater dataset, and both datasets. In 10 out of 12 scenario-dataset combinations, FSL methods achieved at least as good performance as equally powerful baseline models, offering an average improvement of 7\% accuracy points, with up to 9\% and 17\% on optical and sonar, respectively. Further, we found that meta-training on both datasets produced the best performance - an average improvement of 16.5\% and 4.5\% over meta-training on Mini-ImageNet alone and meta-training on underwater dataset alone, respectively. In our experiments, the semi-supervised FSL models performed slightly better than the supervised FSL models offering an average improvement of 4\%. In future work, we plan to reduce some unrealistic assumptions made by FSL methods (e.g., introduce incremental updates) and investigate these methods working alongside an automatic target recognition system to develop a few-shot object detector. 

\section*{Acknowledgement}
We want to give a special thanks to Antti Karjalainen for generating the simulated side-scan sonar data. This work was supported by the EPSRC Centre for Doctoral Training in Robotics and Autonomous Systems, funded by the UK Engineering and Physical Sciences Research Council and SeeByte Ltd (Grant No. EP/S515061/1). 

\bibliographystyle{IEEEtran}  
\bibliography{bib_deep_learning,bib_fsl,bib_datasets,bib_underwater}

\end{document}